\pdfoutput=1

\documentclass[11pt]{article}

\usepackage[final]{acl}

\usepackage{times}
\usepackage{latexsym}

\usepackage[T1]{fontenc}

\usepackage[utf8]{inputenc}

\usepackage{microtype}

\usepackage{pdfpages}
\usepackage[small,compact]{titlesec}
\usepackage[normalem]{ulem}
\useunder{\uline}{\ul}{}
\usepackage{booktabs}       
\usepackage{enumitem}
\usepackage{float}

\newcommand{\norm}[1]{\|#1\|} 

\title{ReFinED: An Efficient Zero-shot-capable Approach to End-to-End Entity Linking}

\author{Tom Ayoola, {\bf Shubhi Tyagi},
        {\bf Joseph Fisher}, {\bf Christos Christodoulopoulos}, {\bf Andrea Pierleoni} \\
        Amazon Alexa AI \\
        Cambridge, UK \\
        \texttt{\{tayoola, tshubhi, fshjos, chrchrs, apierleo\}@amazon.com}}

\begin{document}
\maketitle
\begin{abstract}
We introduce ReFinED, an efficient end-to-end entity linking model which uses fine-grained entity types and entity descriptions to perform linking. The model performs mention detection, fine-grained entity typing, and entity disambiguation for all mentions within a document in a single forward pass, making it more than 60 times faster than competitive existing approaches. ReFinED also surpasses state-of-the-art performance on standard entity linking datasets by an average of 3.7 F1. The model is capable of generalising to large-scale knowledge bases such as Wikidata (which has 15 times more entities than Wikipedia) and of zero-shot entity linking. The combination of speed, accuracy and scale makes ReFinED an effective and cost-efficient system for extracting entities from web-scale datasets, for which the model has been successfully deployed. Our code and pre-trained models are available at \href{https://github.com/alexa/ReFinED}{https://github.com/alexa/ReFinED}.
\end{abstract}

\section{Introduction}
Entity linking (EL) is the task of recognising mentions of entities in unstructured text documents and linking them to the corresponding entities in a Knowledge Base (KB), such as Wikidata. EL is commonly a first stage in systems for question answering \citep{wang-etal-2021-retrieval}, automated KB population \citep{hoffmann-etal-2011-knowledge}, and relation extraction \citep{baldini-soares-etal-2019-matching}.


Currently, EL systems use deep learning methods to learn representations for entities and mentions \citep{ganea-hofmann-2017-deep, le-titov-2018-improving}. Initial techniques learned representations from text alone, which relied on entities appearing in similar contexts in the training data and meant models were only able to link mentions to entities that appeared in the training data. This is problematic both as KBs are continuously growing, and as it is infeasible to build an EL dataset containing all entities in a large KB (such as Wikidata with over 90 million entities). The largest public EL dataset is Wikipedia (using internal hyperlinks as labels), which covers just 3\% of the entities in Wikidata.

Recent models addressed this problem by producing entity representations from a subset of KB information, e.g., entity descriptions \citep{wu-etal-2020-scalable, logeswaran-etal-2019-zero} or fine-grained entity types \citep{DBLP:conf/aaai/OnoeD20, Raiman2018DeepTypeME}, allowing linking to entities not present in the training data or added to the KB after training; termed ``zero-shot'' in the EL literature.\footnote{Note the difference to ``zero-shot'' in the language-model literature, which refers to using no training data for the task.}

However, existing zero-shot-capable EL approaches are an order of magnitude more computationally expensive than non-zero-shot models \citep{rel} as they either require numerous entity types \citep{DBLP:conf/aaai/OnoeD20}, multiple forward passes of a large-scale model to encode mentions and descriptions \citep{wu-etal-2020-scalable}, or regeneration of the input text autoregressively \cite{Cao2020AutoregressiveER}. This makes large-scale processing expensive and thus makes it difficult to benefit from many advantages of zero-shot EL, e.g. the ability to keep up-to-date with new or updated KBs.


In this paper, we propose an efficient end-to-end zero-shot-capable EL model, ReFinED\footnote{ReFinED stands for Representation and Fine-grained typing for Entity Disambiguation.}, which uses fine-grained entity types and entity descriptions to perform entity linking or entity disambiguation (ED; where entity mentions are given). We show that combining information from entity types and descriptions in a simple transformer-based encoder yields performance which is stronger than more complex architectures, surpassing state-of-the-art (SOTA) on 4 ED datasets and 5 EL datasets, and improving overall EL performance by 3.7 F1 points on average across 8 datasets. Importantly, ReFinED performs mention detection, fine-grained entity typing, and entity disambiguation for all mentions within a document in a single forward pass, making it comparable in terms of inference speed to non-zero-shot models. It is 6 times faster than the most efficient zero-shot-capable baseline (which has 9 F1 points lower performance), and more than 60 times faster than more accurate systems (which come within 3 F1 points of ReFinED's average ED performance).



As opposed to previous EL models which primarily use Wikipedia as the target KB, ReFinED targets Wikidata, which enables it to link to 15 times more entities. This is because prior work uses information (e.g. titles, categories, first sentences) from Wikipedia to perform linking. It is unclear whether prior work could be expanded to Wikidata without a drop in performance because entity descriptions are less informative and there are fewer types per entity \citep{weikum2021machine}.

The combination of high accuracy, scalability (with respect to KB size) and fast inference speed makes ReFinED a strong choice for a ``web-scale''\footnote{We refer to corpora with more than 1 billion documents as ``web-scale''.} EL system, in which cost scales approximately linearly with inference speed. We have successfully deployed ReFinED to production in a real-world application and share the lessons learned in Section \ref{deployment-details}.

Our contributions are as follows:
\vspace{-2mm}
\begin{enumerate}
	\item{We build a simple and efficient zero-shot capable end-to-end EL model using entity descriptions and entity typing, which outperforms previous approaches on standard-EL datasets by 3.7 F1 points on average.}
	\vspace{-2mm}
	\item{We demonstrate our model is more than 6 times faster than existing low-accuracy zero-shot capable systems, and 60 times faster than higher-accuracy systems, whilst also being capable of disambiguating against Wikidata-scale entity sets. The combination of accuracy, speed and scale makes the model suitable for web-scale information extraction.} 
	\vspace{-2mm}
	\item{We release our code and models.}
\end{enumerate}

\section{Related work}

\paragraph{Single architecture for entity linking}
EL consists of two main tasks, mention detection (MD) and ED. MD involves recognising mentions of entities in text, and ED assigns a KB entity to each mention. We follow \citep{kolitsas-etal-2018-end} in training a joint model for MD and ED. 


\paragraph{Entity disambiguation with fine-grained entity typing}
In \citet{DBLP:conf/aaai/OnoeD20} and \citet{Raiman2018DeepTypeME} ED is formulated as an entity typing problem. A fine-grained entity typing model is trained on a distantly-supervised dataset consisting of over 10k types derived from Wikipedia categories (e.g. movies released in a specific year). The entity typing model is then used to link entities. We extend their approach to Wikidata, by using a subset of Wikidata triples for providing types instead of Wikipedia categories. 

\paragraph{Entity disambiguation with entity descriptions}
Several recent works have used entity descriptions for ED \citep{wu-etal-2020-scalable, logeswaran-etal-2019-zero}. Typically, descriptions are sourced from Wikipedia by joining the entity's title with the first sentence of the Wikipedia article. Entities are ranked by concatenating mention context and entity description, then passing each mention-entity pair to a cross-encoder. \citet{wu-etal-2020-scalable} shows a cross-encoder outperforms a bi-encoder, with the latter missing many fine-grained interactions between context and description. In our work, we find that a bi-encoder is sufficient to achieve SOTA performance when combined with fine-grained entity typing, and generalise the approach from Wikipedia (6M entities) to Wikidata (90M entities).\footnote{We replace Wikipedia titles with Wikidata labels, and Wikipedia sentences with Wikidata entity descriptions.}

\section{Proposed method}
\subsection{Task Formulation}
Given a KB\footnote{We assume entities in the KB have a textual description and a collection of facts.} with a set of entities $E = \{e_1, e_2, \dots, e_{|E|}\}$, let $X = [x_1, x_2, \dots, x_{|X|}]$ be a sequence of tokens in the document, and $M = \{m_1, m_2, ... m_{|M|}\}$ be a set of entity mentions. The goal of ED is to create a function $\mathcal{M}: M \rightarrow E$ which assigns each mention the correct entity label. In EL, both the mention spans and entity labels need to be predicted. We only consider mentions with a valid gold entity in the KB during evaluation.

\subsection{Overview}
We propose an end-to-end EL model which is jointly optimised for mention detection, fine-grained entity typing, and entity disambiguation for all mentions within a document in a single forward pass. In this section, we describe the components of our model, depicted in Figure \ref{model_diagram}.

\begin{small}
\begin{figure*}[h]
	\centering
	\hspace{-25pt}
	\includegraphics[width=380pt]{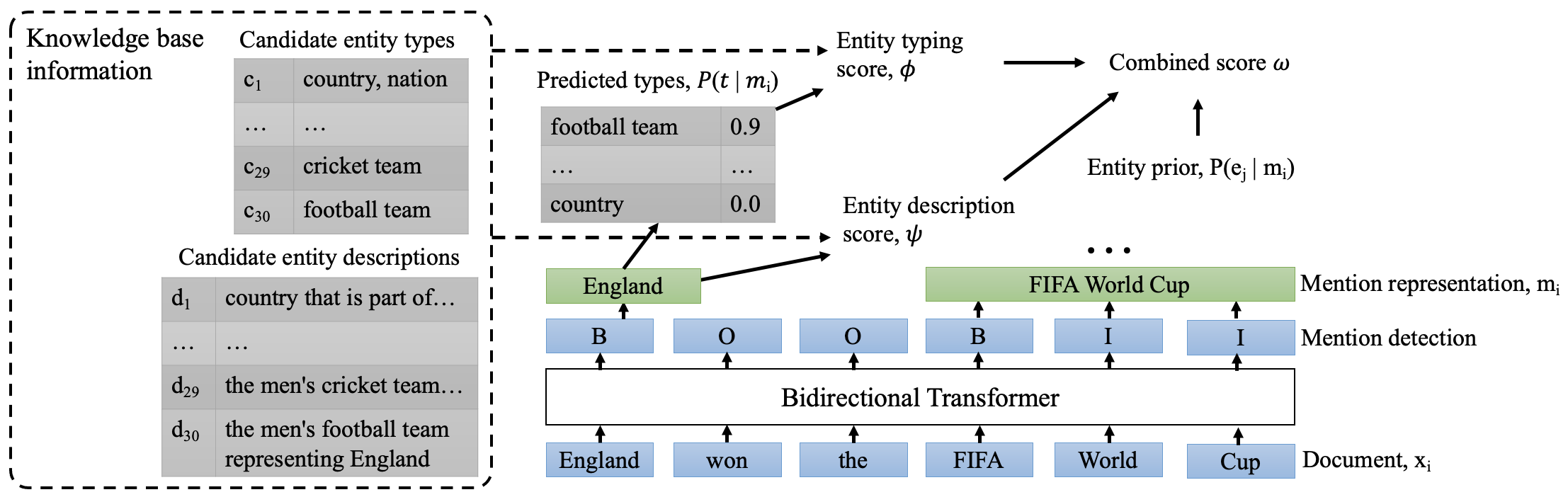}  
	\caption{Our model architecture shown for a document with two mentions, \textit{England} and \textit{FIFA World Cup}. The model performs mention detection, entity typing, and entity disambiguation for all mentions in a single pass.}
	\label{model_diagram}
\end{figure*}
\end{small}

\subsection{Context representation}
We encode the tokens $x_i$ in the input text document using a Transformer model. We use the contextualised token embeddings from the final layer, denoted as $\mathbf{h_i}$ for the token $x_i$.\footnote{We use bold letters for vectors throughout our paper, and treat $m_i$ and $\mathbf{m_i}$ as different terms.}

\subsection{Mention detection}
Entity linking requires entity mentions to be predicted. We encode mentions using the BIO tagging format \citep{ramshaw-marcus-1995-text} with 3 labels which indicate whether a token is at the beginning, inside of, or outside of a mention. We train a linear layer to perform token classification from the contextualised token embeddings $\mathbf{h_i}$ using cross-entropy loss $\mathcal{L}_m$ with respect to the gold token labels.

\subsection{Mention representation}\label{mention representation}
A fixed-length embedding $\mathbf{m_i}$ for each mention $m_i$ is obtained by average pooling the contextualised tokens embeddings of the mention. All mentions $M$ in a document $X$ are encoded in a single forward pass, which improves efficiency relative to previous work that require a forward-pass for each mention \cite{wu-etal-2020-scalable, Orr2021BootlegCT}.
\subsection{Entity typing score $\mathbf{\phi}$}
Given a fixed set of types $t \in T$ from a KB, where $t$ is a relation-object pair $(r, o)$ (e.g. (instance of, song)), we predict an independent probability for each type $t$ for each mention by applying a linear layer $f_1$ followed by a sigmoid activation to the mention embedding $\mathbf{m_i}$. To score mention-entity pairs using predicted types, we calculate the Euclidean distance (L2 norm) between predicted types and the candidate entity's types $\mathbf{c_j}$ binary vector\footnote{We use 1 to indicate the presence of an entity type and 0 the absence of an entity type for our binary vector. Note that a single entity can have multiple entity types.}:
\begin{equation}
 \phi(e_j, m_i) = \norm{\sigma(f_1(\mathbf{m_i})) - \mathbf{c_j}}_2
\end{equation}
We follow \citet{DBLP:conf/aaai/OnoeD20} by training the entity typing module on distantly-supervised type labels from the gold entity using binary cross-entropy loss $\mathcal{L}_{t}$. See Appendix \ref{sec:type-selection-details} for details on the choice of types $T$.

\subsection{Entity description score  $\mathbf{\psi}$} \label{descriptionscore}
We use a bi-encoder architecture similar to the work of \citet{wu-etal-2020-scalable} but modified to encode all mentions $m_i$ in a document simultaneously (as explained in Section \ref{mention representation}). We represent KB entities as:
\vspace{-4mm}
\begin{itemize}
	\item[] [CLS] label [SEP] description [SEP]
\end{itemize}
\vspace{-2mm}
where label and description are the tokens of the entity label and entity description in the KB. We use a separate Transformer model (trained jointly with our mention transformer) to encode the representation of KB entities $e_j$ into fixed dimension vectors (description embeddings) $\mathbf{d_j}$ by taking final layer embedding for the [CLS] token. We apply linear layers $f_2$ and $f_3$ to the mention embeddings $\mathbf{m_i}$ and entity description embeddings $\mathbf{d_j}$ respectively to project them to a shared vector space. We calculate the dot product between the two projected embeddings to compute the entity scores:
\begin{equation}
	 \psi(e_j, m_i) = f_2(\mathbf{m_i}) \cdot f_3(\mathbf{d_j})
\end{equation}
We train this module using cross-entropy loss $\mathcal{L}_d$, with respect to gold entity label.

\subsection{Combined score $\mathbf{\omega}$}\label{combined_score}
We compute a combined score $\omega$ by applying a linear layer (with output dimension 1) $f_4$ on top of the concatenation of entity typing score, entity description score, and a global entity prior $\hat{P}(e|m)$. The global entity prior is obtained from a corpus \citep{hoffart-etal-2011-robust} or a popularity metric \citep{diefenbach:hal-01905724}. We include $\hat{P}(e|m)$ to improve results for cases where context is limited (e.g. short question text). In addition, we add a special candidate for the NIL entity with an unnormalised score of 0, which indicates none of the candidate entities are correct.
\begin{equation}
	\omega(e_j, m_i) = f_4(\psi(\mathbf{e_j}, \mathbf{m_i}); \phi(\mathbf{e_j}, \mathbf{m_i}); \hat{P}(\mathbf{e_j}|\mathbf{m_i}))
\end{equation}
We train this module using cross-entropy loss $\mathcal{L}_c$ with respect to the gold entity label.
\subsection{Optimisation and inference}
We optimise the model using a weighted sum of the module-specific losses with fixed weights, which are tunable hyperparameters. At training time, we use the provided mention spans instead of the predicted mention spans and train mention detection alongside the other tasks:
\begin{equation}
	\mathcal{L} = \lambda_1 \mathcal{L}_m + \lambda_2 \mathcal{L}_t + \lambda_3 \mathcal{L}_d +  \lambda_4 \mathcal{L}_c
\end{equation}

For EL inference, we use the predicted mention spans and take the KB entity (or NIL) with the highest combined score.
For ED inference, we use the provided gold mention spans.

\subsection{Zero-shot ED}
Our proposed method is able to link to zero-shot (unseen during training) entities because it scores entities based on types and descriptions. New entities can be introduced by updating entity lookups.

\section{Experiments} \label{experiments}





\subsection{Entity disambiguation}
\paragraph{Non-zero-shot ED} \label{ed_setup}
We evaluate our model on the ED task using the same experimental setting as previous work \citep{ganea-hofmann-2017-deep, le-titov-2018-improving, Cao2020AutoregressiveER}. We pretrain on Wikipedia, then use the AIDA-CoNLL dataset \citep{hoffart-etal-2011-robust} to fine-tune and evaluate. We measure out-of-domain performance on the datasets MSNBC \citep{msnbc}, AQUAINT \citep{aquaint}, ACE2004 \citep{ace2004}, WNED-CWEB (CWEB) \citep{cweb} and WNED-WIKI (WIKI) \citep{wned}. 
We report \emph{InKB} micro-F1 \cite{gerbil}. We also evaluate on AIDA-CoNLL using the candidate list generated by \citet{pershina-etal-2015-personalized}, known as PPRforNED, for the sake of comparison with previous SOTA results.


\paragraph{Zero-shot ED}
To compare our method to previous work, we measure zero-shot ED performance using the WikiLinksNED Unseen Mentions dataset \citep{wikilinksned, DBLP:conf/aaai/OnoeD20}. The dataset contains a diverse set of ambiguous entities spanning multiple domains. We train our model on the provided training data and evaluate accuracy on the test set for seen and unseen (zero-shot) entities.

\subsection{Entity linking}
\paragraph{Non-zero-shot EL}
Following previous work \citep{kolitsas-etal-2018-end, Cao2020AutoregressiveER}, we use the GERBIL platform \citep{gerbil} to evaluate EL. We evaluate \emph{InKB} micro-F1 with strong matching (predictions must match exactly the gold mention boundaries). Similarly to the non-zero-shot ED experiment, we pretrain on Wikipedia, then use the AIDA-CoNLL dataset for fine-tuning and evaluation. For out-of-domain performance evaluation we use MSNBC \citep{msnbc}, OKE-2015, OKE-2016 \citep{oke}, N3-Reuters-128 (R128), N3-RSS-500 \citep{rss500}, Derczynski \citep{derczynski}, KORE50 \citep{KORE50}.


\subsection{Inference speed}
We compare the computational efficiency of our model to three high-performing EL systems \citep{wu-etal-2020-scalable, Cao2020AutoregressiveER, Orr2021BootlegCT} for which code is available. We benchmark both modes of BLINK \citep{wu-etal-2020-scalable}; the bi-encoder (encodes mention and entities independently) and the more accurate cross-encoder (encodes mention and entities jointly).\footnote{We use a max context length of 128 tokens and pre-computed entity embeddings for the bi-encoder. For the cross-encoder, we use max context length of 32 tokens.} We measure the time to perform end-to-end EL inference on the AIDA-CoNLL test dataset using a single V100 GPU. The dataset consists of 231 documents and 4464 mentions.

\subsection{Training details}

\begin{table*}[h]
	\centering
 	\resizebox{.82\linewidth}{!}{%
		\begin{tabular}{@{}llccccccc@{}}
			\toprule
			& \textbf{Method}               & \textbf{AIDA}                 & \textbf{MSNBC*} & \textbf{AQUAINT*} & \textbf{ACE2004*} & \textbf{CWEB*} & \multicolumn{1}{c|}{\textbf{WIKI*}} & \textbf{Avg.}        \\ \midrule
			& \citet{yang-etal-2018-collective}           & {93.0}                    & 92.6           & 89.9             & 88.5             & {\ul 81.8} & \multicolumn{1}{c|}{79.2}          & 87.5                   \\
			& \citet{yang-etal-2019-learning}                                     & {93.7} & 93.8                & 88.3                
& {90.1}          & 75.6 & \multicolumn{1}{c|}{78.8}                & 86.7          \\
			& \citet{fang}            & \textbf{94.3}                          & 92.8           & 87.5             & {\ul 91.2}       & 78.5          & \multicolumn{1}{c|}{82.8}          & 87.9                 \\
			& \citet{wu-etal-2020-scalable}$^{\dag}$            & 86.7                          & 90.3           & 88.9             & 88.7       & \textbf{82.6}          & \multicolumn{1}{c|}{86.1}          & 87.2                 \\
			& \citet{Cao2020AutoregressiveER}                  & 93.3                          & {\ul 94.3}     & 89.9             & 90.1             & 77.3          & \multicolumn{1}{c|}{{\ul 87.4}}    & {\ul 88.7}           \\ 
			& \citet{Orr2021BootlegCT}$^{**}$                                         & 80.9           & 80.5                & 74.2                & 83.6                & 70.2             & \multicolumn{1}{c|}{76.2}                   & 77.6             \\ \midrule
			& \textbf{ReFinED (Wikipedia)}  & 87.5                          & \textbf{94.4}  & \textbf{91.8}    & \textbf{91.6}    & 77.8          & \multicolumn{1}{c|}{\textbf{88.7}} & 88.6                 \\
			& \textbf{ReFinED (fine-tuned)} & {\ul 93.9}                          & 94.1           & {\ul 90.8}       & 90.8             & 79.4    & \multicolumn{1}{c|}{{\ul 87.4}}    & \textbf{89.4}        \\
            \midrule  \midrule
			Ablations & w/o entity priors (Wikipedia)             & 86.3                          & 93.7           & 86.0             & 92.8             & 76.0          & \multicolumn{1}{c|}{88.3}          & 87.2                 \\
            & w/o entity types (Wikipedia)              & 82.2                          & 92.6           & 91.1             & 90.1             & 76.5          & \multicolumn{1}{c|}{87.0}          & 86.6                 \\
            & w/o descriptions (Wikipedia)              & 85.7                          & 93.9           & 89.5             & 91.2             & 76.1          & \multicolumn{1}{c|}{84.3}          & 86.8                 \\
            & w/o pretraining (fine-tuned)               & 88.2                          & 92.3           & 86.8             & 90.6             & 75.1          & \multicolumn{1}{c|}{74.5}          & 84.6                \\
        \bottomrule
		\end{tabular}
		}
		\caption{ED InKB micro F1 scores on in-domain and out-of-domain test sets. The best value in \textbf{bold} and second best is {\ul underlined}. $^{\dag}$Normalised accuracy is reported. *Out-of-domain datasets. $^{**}$Result obtained using code released by authors.}
     	\label{ed_results}
\end{table*}

\paragraph{Candidate generation}\label{candidate_gen}
We follow \citet{le-titov-2018-improving} by selecting the top-30 candidate entities using entity priors.\footnote{Derived from Wikipedia hyperlink count statistics, YAGO, a large Web corpus and Wikidata aliases.} For training, we only keep 5 candidates, 1 gold candidate, 2 candidates with the highest $\hat{p}(e_j|m_i)$ and 2 random candidates. When the gold entity is not in the candidate list during training, we use NIL as the correct label.

\paragraph{Wikipedia pretraining} \label{Wikipedia}
We use Wikidata as our KB (i.e. for entity types and descriptions). To make comparisons reliable, we restrict to the set of entities in English Wikipedia (total of 6.2M). We build a training dataset from the 2021-02-01 dump of Wikipedia and Wikidata and use hyperlinks as entity labels. To increase entity label coverage, we add weak labels to mentions of the article entity \citep{Orr2021BootlegCT, broscheit-2019-investigating, Cao2020AutoregressiveER}.\footnote{We add weak labels by using simple heuristics such as matching mentions to the page's title.} The dataset consists of approximately 100M mention-entity pairs. We use entity labels to generate entity type labels, as in \citet{DBLP:conf/aaai/OnoeD20}. In addition, we follow \citet{Fvry2020EmpiricalEO} by adding mention labels to unlinked mentions using a named entity recogniser to provide additional mention detection signal.

\paragraph{Model details}
We divide the documents into chunks of 300 tokens and subsample 40 mentions per chunk during pretraining. The model is trained for 2 epochs on Wikipedia and the transformers are initialised with RoBERTa \citep{roberta} base weights. The description transformer has 2 layers. BERT-style masking \citep{devlin-etal-2019-bert} is applied to mentions during pretraining. During fine-tuning and evaluation, we increase the sequence length to 512 and set the maximum candidate entities to 30.

\section{Results}


\subsection{Entity disambiguation}
\paragraph{Non-zero-shot ED} \label{ed_results_para}
We report \emph{InKB} micro-F1 (with and without fine-tuning on AIDA) and compare it with SOTA ED models in Table \ref{ed_results}. Our model performs strongly across all datasets, surpassing the previous average F1 across the 6 datasets by 0.7 F1 points. We observe the model achieves SOTA performance on 4 out of the 6 datasets without fine-tuning, suggesting it is able to learn patterns from Wikipedia that transfer well to other domains. Nonetheless, fine-tuning on the AIDA-CoNLL dataset leads to a substantial improvement (+6.4 F1 points) which can be attributed to the model learning peculiarities of the dataset (e.g. cricket score tables).

The ablations in Table \ref{ed_results} show entity types and entity descriptions are complementary (+2.0 F1 points when combined). This is explained by increased robustness to partially missing entity information (e.g. KB entities without descriptions) and different knowledge being expressed. Entity priors are useful but contribute less than other components of our combined score (Section \ref{combined_score}). Without priors, F1 falls by 5.0 points on AQUAINT and increases by 1.2 points on ACE2004, which is expected as AQUAINT contains a high proportion of popular entities, and ACE2004 more rare entities. Pretraining has the largest impact on ED performance, particularly on datasets such as WIKI (+12.0 F1) derived from encyclopedia text.
\begin{table}[h]
\centering

\label{ppr_results}
\resizebox{.55\linewidth}{!}{%

\begin{tabular}{@{}lc@{}}
		\toprule
		\textbf{Method}      & \textbf{AIDA} \\ \midrule
		\citet{DBLP:conf/aaai/OnoeD20}                & 85.9          \\
		\citet{Raiman2018DeepTypeME}              & 94.9          \\
		\citet{Orr2021BootlegCT}                & {\ul 96.8}    \\ \midrule
		\textbf{ReFinED (Wikipedia)}  & 89.1          \\
		\textbf{ReFinED (fine-tuned)} & \textbf{97.1} \\ \bottomrule
\end{tabular}
}
\caption{ED accuracy on AIDA-CoNLL using PPRForNED candidates.}
\label{ppr_results}
\end{table}

\begin{table*}[h]
	\centering
 	\resizebox{.95\linewidth}{!}{
		\begin{tabular}{@{}lccccccccc@{}}
			\toprule
			\textbf{Method}         & \textbf{AIDA}                 & \textbf{MSNBC*} & \textbf{DER*}  & \textbf{K50*}  & \textbf{R128*} & \textbf{R500*} & \textbf{OKE15*} & \multicolumn{1}{c|}{\textbf{OKE16*}} & \textbf{Avg.}        \\ \midrule
			\citet{hoffart-etal-2011-robust}   & 72.8                          & 65.1           & 32.6          & 55.4          & 46.4          & \textbf{42.4} & 63.1           & \multicolumn{1}{c|}{0.0}              & 47.2                 \\
			\citet{kolitsas-etal-2018-end}   & 82.4                          & {\ul 72.4}     & 34.1          & 35.2          & 50.3    & 38.2          & 61.9           & \multicolumn{1}{c|}{52.7}           & 53.4                 \\
			\citet{rel} & 80.5                          & {\ul 72.4}     & 41.1          & 50.7          & 49.9          & 35.0            & 63.1           & \multicolumn{1}{c|}{58.3}           & 56.4                 \\
			\citet{Cao2020AutoregressiveER}            & {\ul 83.7}                    & \textbf{73.7}  & \textbf{54.1} & 60.7          & 46.7          & 40.3          & 56.0             & \multicolumn{1}{c|}{50.0}             & 58.2                 \\ \midrule
			\textbf{ReFinED (Wikipedia)} & 77.8                          & 70.0           & 49.0          & \textbf{65.9} & {\ul 52.6} & 40.1          & \textbf{65.0}    & \multicolumn{1}{c|}{\textbf{59.5}}  & {\ul 60.0}           \\
			\textbf{ReFinED (fine-tuned)}   & \textbf{84.0}                 & 71.8           & {\ul 50.7}    & {\ul 64.7}    & \textbf{58.1} & {\ul 42.0}    & {\ul 64.4}     & \multicolumn{1}{c|}{{\ul 59.1}}     & \textbf{61.9}        \\ \bottomrule
		\end{tabular}
		}
		\caption{EL InKB micro F1 scores on in-domain and out-of-domain test sets reported by Gerbil. The best value in \textbf{bold} and second best is {\ul underlined}. *Out-of-domain datasets.}
     	\label{el_results}
\end{table*}
Table \ref{ppr_results} shows accuracy on the AIDA-CoNLL dataset when we use PPRforNED candidates. ReFinED outperforms purely entity typing approaches \citep{Raiman2018DeepTypeME, DBLP:conf/aaai/OnoeD20} by a margin of +2.2\% accuracy, due to the addition of entity descriptions.

\paragraph{Zero-shot ED}
In Table \ref{wikilinksned_results}, we report ED accuracy on the WikiLinksNED Unseen Mentions test set for seen and unseen entities. Our model outperforms the baseline by 3.0 F1, with, surprisingly, 6.6\% higher accuracy for unseen than for seen entities. We find this is partly due to higher top 30 candidate recall for the unseen entity subset (95.0\% compared to 91.1\% for the seen entity subset) and also because our mention masking strategy reduces the reliance of entities appearing in the training data with similar surface forms. Moreover, ReFinED uses entity types and descriptions to link entities instead of relying on entity memorisation, which means the number of training examples for a given entity will not necessarily correlate with performance. The number of similar entities in the training dataset and the ambiguity of the test examples \cite{provatorova-etal-2021-robustness} will likely have more significant influence on performance.


\begin{table}[h]
\centering
\resizebox{.80\linewidth}{!}{%

\begin{tabular}{@{}lcc|c@{}}
		\toprule
		\textbf{Method}  & \textbf{Seen} & \textbf{Unseen} & \textbf{Total} \\ \midrule
		\citet{Cao2020AutoregressiveER}            & 64.3          & 63.2            & 63.5           \\ \midrule
		\textbf{ReFinED} & 61.6          & 68.2            & 66.5           \\ \bottomrule
	\end{tabular}
	}
	\caption{ED accuracy on WikiLinksNED Unseen Mentions test.}
	\label{wikilinksned_results}
\end{table}

\subsection{Entity linking}
EL results are shown in Table \ref{el_results}. ReFinED outperforms other models on all but 3 datasets, often by a considerable F1 point margin (e.g. 7.8 on N3-Reuters-128 and 4.0 on KORE50) and improves the average across all 8 datasets by 3.7 F1 points.
EL improves as ED and mention detection can generalise to different datasets due to the model being pretrained on Wikipedia hyperlinks as opposed to only AIDA-CoNLL. We also report results on the ISTEX and WebQSP datasets in Appendix \ref{sec:additional_results}.

\subsection{Inference speed} \label{speed_test}

Table \ref{tab:inference-speed-test} shows the time taken to run inference on the AIDA-CoNLL test dataset, alongside the average ED performance. ReFinED is 6 times faster than the BLINK \cite{wu-etal-2020-scalable} bi-encoder, which also has an average F1 which is 9 points lower. Compared to the higher accuracy systems, ReFinED is 60 times faster than the BLINK cross-encoder, and 140 times faster than the autoregressive approach of \citet{Cao2020AutoregressiveER}. This is because ReFinED uses a single forward pass to jointly encode all mentions and candidate KB entities in the document (512 token chunk), and hence requires $\approx$ 231 forward passes for the full dataset. The bi-encoder model requires $\approx$ 4464 forward passes as mentions are encoded individually, and the cross-encoder baseline requires $\approx$ 90k forward passes as each mention is encoded with each candidate. The autoregressive approach suffers from high computational cost due to the deep decoder, which generates a single token at a time. Also, all baselines require a separate model for MD whereas ReFinED performs end-to-end EL using a single model, which improves efficiency and simplifies model deployment.

\begin{table}[h]
    
	\centering
	\resizebox{.99\linewidth}{!}{%
	\begin{tabular}{@{}lcc@{}}
		\toprule
		\textbf{Method}     & \textbf{Time taken (s)} & \textbf{Avg. ED F1}\\ \midrule
		\citet{Cao2020AutoregressiveER} & 2100 & 88.7 \\
		\citet{wu-etal-2020-scalable} bi-encoder    &   93  & 80.4 \\
		\citet{wu-etal-2020-scalable} cross-encoder & 917   & 87.2                  \\ 
		\citet{Orr2021BootlegCT} & 438 & 77.6 \\ 
		\midrule
		\textbf{ReFinED}    & \textbf{15} & \textbf{89.4}           \\ \bottomrule
	\end{tabular}
	}
	\caption{Time taken in seconds for EL inference on AIDA-CoNLL test dataset.}\label{time_taken}
	\label{tab:inference-speed-test}

\end{table}

\section{Deployment Details}\label{deployment-details}

We have successfully deployed the ReFinED EL model in a real-world application, the aim of which is to populate a KB by extracting facts from unstructured text found on web pages with high precision. The application requires running ReFinED on a billion web pages (in which we link 25 billion mentions) multiple times per year. The scale of this deployment highlighted a number of learning points. 


Firstly, the entity linking model must be computationally efficient. The inference speed of ReFinED allows the processing of the billion web pages in 27k machine hours (2 days using 500 instances), on machines with a single T4 GPU. Given availability of cloud compute, the cost of processing the same documents with the models evaluated in Section \ref{speed_test} would scale approximately linearly with their inference speeds. That is, the BLINK bi-encoder would require 3000 instances for 2 days, or 500 instances for 12 days, implying a roughly 6-fold increase in cost. 

Secondly, the scale of the number of pages also brings with it diversity of domains, meaning the model benefits from linking to a large catalogue of entities (over 90 million) - including zero-shot entities.

Thirdly, we observed that deploying an end-to-end self-contained EL model is easier to horizontally scale and has a lower operational cost than deploying multiple systems for each subcomponent (such as candidate generation). 

Finally, in real-world data, unlike in ED datasets, there are a large number of cases where the correct entity does not exist in the KB. This meant that we had to train the model on examples where the correct entity was not in the candidate list to reduce overprediction.









\section{Conclusion}

We propose a scalable end-to-end EL model which uses entity types and entity descriptions to perform linking. Our model achieves SOTA results for both ED (+0.7 F1 points on average across 6 datasets) and EL (+3.7 F1 points on average across 8 datasets) while being 60 times faster than comparatively accurate baselines. We demonstrate our approach scales well to a KB (Wikidata) 15 times larger than Wikipedia while maintaining competitive performance. The combination of accuracy, speed and scale means the system is capable of being deployed to extract entities from web-scale datasets with higher accuracy and an order of magnitude lower cost than existing approaches. 


\section*{Acknowledgements}
The authors would like to thank Clara Vania, Grace Lee, and Amir Saffari for helpful discussions and feedback. We also thank the anonymous reviewers for valuable comments that improved the quality of the paper.

\clearpage

\bibliography{anthology,custom}
\bibliographystyle{acl_natbib}

\clearpage

\appendix

\section{Entity Type Selection}
\label{sec:type-selection-details}
Our entity types are formed from Wikidata relation-object pairs and relation-object pairs inferred from the Wikidata subclass hierarchy (for example, (instance of, organisation) can be inferred from (instance of, business)). We only consider types with the following relations: instance of, occupation, country, sport. We select types by iteratively adding types that separate (assuming an oracle type classifier) the gold entity from negative candidates for the most examples in our Wikipedia training dataset.

Type information stored in KBs often varies in granularity between entities (e.g. some capital city entities have the type capital city and others only city),  adversely affecting training signal. To mitigate this, we use the class hierarchy to add parent types to entities. This injects class hierarchy information into the model, enabling type granularity to depend on context. 


\section{Training Details}
\label{sec:training-details}
We use the Hugging Face implementation of RoBERTa \cite{huggingface} and optimise our model using Adam \cite{adam} with a linear learning rate schedule. We ignore the loss from mentions where the gold entity is not in the candidate set. The named-entity recogniser, used to preprocess our Wikipedia training dataset, is a RoBERTa token classification model trained on the AIDA-CoNLL dataset mention boundaries. We add weak  entity labels for mentions that match the page's title (or surname for Wikipedia pages about people). We present our main hyperparameters in Table \ref{tab:hyperparams}. Due to the high computational cost of training the model, we did not conduct an extensive hyperparameter search. Training on Wikipedia took approximately 48 hours on a single machine with 4 V100 GPUs. The model has approximately 154M parameters (123 million in the roberta-base architecture, and 31M for the additional description encoder and output layers). 

\begin{table}[h]
\resizebox{0.9\linewidth}{!}{%
\begin{tabular}{@{}ll@{}}
\toprule
\textbf{Hyperparameter} & \textbf{Value} \\ \midrule
learning rate           & 3e-5           \\
batch size              & 64             \\
max sequence length     & 300            \\
dropout                 & 0.05           \\
description embeddings dim.    & 300            \\ 
\# training steps                & 1M             \\
\# candidates           & 30             \\
\# entity types      & 1400            \\ 
mention transformer init.  & roberta-base \\
\# mention encoder layers     & 12            \\
description transformer init.  & roberta-base \\
\# description encoder layers     & 2            \\ 
\# description tokens     & 32            \\
$\lambda_1$, $\lambda_2$, $\lambda_3$, $\lambda_4$      & (0.01, 1, 0.01, 1)            \\
mention mask prob.      & 0.7            \\

\bottomrule
\end{tabular}
}
\caption{Our model hyperparameters}
\label{tab:hyperparams}
\end{table}

\section{Additional results}
\label{sec:additional_results}

\paragraph{Wikidata ED experimental setup}
To measure ED performance on non-Wikipedia entities, we expand our entity set to Wikidata (which has over 90M entities) and evaluate our model on the ISTEX test dataset \citep{delpeuch2020opentapioca}. We add labels and aliases from Wikidata for candidate generation and remove entity priors from our entity scoring (Section \ref{combined_score}). 

\paragraph{Wikidata ED results}
We evaluate ED performance on the ISTEX dataset (which targets Wikidata). Our model outperforms \citet{delpeuch2020opentapioca} (92.1 vs 87.0 micro F1) which uses hand-crafted features specifically designed for linking Wikidata entities. This shows that our approach scales to Wikidata and generalises well when there is increased mention ambiguity. Our model performs 0.5 F1 points below the SOTA \citet{Mulang__2020} (92.6 vs 92.1 micro F1) which is likely due to differing candidate entity generation methods.

	

\paragraph{Entity Linking performance on questions}
We report results on the WebQSP dataset in Table \ref{webqsp_results}, which shows EL performance on questions. Our model has similar performance to ELQ, which is SOTA on WebQSP and is optimised for questions. Our model is faster than all baselines which can be attributed to using an end-to-end EL model, restricting ED predictions to the predicted mentions only, and using a smaller model (compared to ELQ which uses BERT-large \citep{devlin-etal-2019-bert}). 

\begin{table}[H]

\centering
\resizebox{\linewidth}{!}{%
\begin{tabular}{@{}lcc@{}}
\toprule
\textbf{Method}               & \textbf{WebQSP} & \textbf{\#Q/s} \\ \midrule
TAGME \citep{tagme}                         & 36.1            & {\ul 2.39}                          \\
BLINK \citep{wu-etal-2020-scalable} (Wikipedia)             & 80.8            & 0.80                          \\
ELQ \citep{li-etal-2020-efficient} (Wikipedia)               & 83.9            & 1.56                          \\
ELQ \citep{li-etal-2020-efficient} (fine-tuned)              & {\ul 89.0}      & 1.56                          \\ \midrule
\textbf{ReFinED (Wikipedia)}  & 84.1            & \textbf{2.78}                             \\
\textbf{ReFinED (fine-tuned)} & \textbf{89.1}   & \textbf{2.78}                             \\ \bottomrule
\end{tabular}
}
\caption{Entity linking weak matching InKB micro F1 scores on WebQSP EL dataset \citep{li-etal-2020-efficient}. \#Q/s is number of questions per second for a single CPU.}
\label{webqsp_results}
\end{table}

\section{Dataset statistics}

We present the topic, number of documents and number of mentions for each dataset used for evaluation. The datasets used cover a variety of sources including wikipedia text, news articles, web text and tweets. Note that the performance of the model outside these domains may be significantly different. 

Note also that all datasets used are for English only, allowing comparison to previous work. Our method is extendable to any language for which there is an language-specific version of Wikipedia on which the model could be trained. However, we cannot guarantee the accuracy of the model across these languages without further experimentation. 

\begin{table}[h]
	\centering
 	\resizebox{\linewidth}{!}{
		\begin{tabular}{lccc}
		& \textbf{Topic} & \textbf{Num docs} & \textbf{Num Mentions} \\
		\toprule
		\textbf{AIDA} & news & 231 & 4464 \\
		\textbf{MSNBC} & news & 20 & 656\\
		\textbf{AQUAINT} & news & 50 & 743 \\
		\textbf{ACE2004} & news & 57 & 259\\
		\textbf{CWEB} & web & 320 & 11154\\ 
		\textbf{WIKI} & Wikipedia & 320 & 6821\\
		\textbf{WikilinksNED} & web & 10000 & 10000\\
        \bottomrule
		\end{tabular}
		}
		\caption{Dataset statistics for entity disambiguation datasets}
     	\label{dataset_statistics_ed}
\end{table}

\begin{table}[h]
	\centering
 	\resizebox{\linewidth}{!}{
		\begin{tabular}{lccc}
		& \textbf{Topic} & \textbf{Num docs} & \textbf{Num Mentions} \\
		\toprule
		\textbf{AIDA} & news & 231 & 4464 \\
		\textbf{MSNBC} & news & 20 & 656 \\
		\textbf{DER} & tweets & 182 & 242 \\
		\textbf{K50} & mixed & 50 & 145\\
		\textbf{R128} & news & 128 & 638 \\ 
		\textbf{R500} & news & 500 & 530 \\
		\textbf{OKE15} & Wikipedia & 199 & 1017\\
		\textbf{OKE16} & Wikipedia & 254 & 1402\\
        \bottomrule
		\end{tabular}
		}
		\caption{Dataset statistics for entity linking datasets}
     	\label{dataset_statistics_el}
\end{table}

\end{document}